%% file: main.tex
\documentclass[letterpaper]{article}
\usepackage[draft]{aaai2026}
\usepackage{times}
\usepackage{helvet}
\usepackage{courier}

\usepackage[hyphens]{url}
\usepackage{graphicx}
\usepackage{amsmath}
\usepackage{natbib}
\usepackage{caption}
\usepackage{xspace}
\newcommand{\method}{\textsc{QCR}\xspace}

\title{Beyond Retrieval: Query-Conditioned Reuse of Long-Horizon Agent Trajectories}
\author{
Yifei Li$^{1,2,*}$,
Heng Wang$^{1,2,*}$,
Lingling Zhang$^{1,2,\dagger}$,
Muye Huang$^{1,2}$,
Xinyu Zhang$^{1,2}$,
Jiashuai Liu$^{1,2}$,
Hang Yan$^{1,2}$,
Rongman Xu$^{1,2}$
}

\affiliations{
$^{1}$School of Computer Science and Technology, Xi'an Jiaotong University\\
$^{2}$MOE KLNN Lab, Xi'an Jiaotong University\\
$^{*}$Equal contribution. \quad $^{\dagger}$Corresponding author.\\
\texttt{yifeilee@stu.xjtu.edu.cn}
}

\begin{document}
\maketitle

\begin{abstract}
Retrieval can identify a past trajectory that may matter, yet it does not specify how an acting agent should use that trajectory after users, entities, constraints, or environment state have changed. We identify this post-retrieval reuse step as a distinct bottleneck for long-horizon trajectory memory and formulate an evaluation framework that holds candidate retrieval, target state, model, decoding, and tool budget fixed while varying the support delivered to the agent. We instantiate the framework with query-conditioned reuse (QCR), a deliberately simple target-bound note with a workflow invariant, bindings to re-obtain, applicability conditions, and a verification guardrail. QCR serves to test the reuse hypothesis rather than to claim a universally preferred memory format. Across 2,391 target instances in WebArena, WorkArena, and AppWorld, QCR reaches 62.3\% average Success, 10.7 points above Full Trajectory, while using 48.9\% fewer online tokens. Summary reranking selects a reusable memory for 94.8\% of targets, placing end-task Success within 1.8 points of an oracle reusable selector. Analyses by trajectory length and source--target binding shift show that direct trajectory injection loses much of its utility as traces grow longer or source-specific values change, whereas target-bound support preserves a larger share of the measured gain. The resulting framework separates retrieval quality from the problem of turning retrieved experience into safe, useful support for a new task.
\end{abstract}

\section{Introduction}
\begin{figure}[t]
\centering
\includegraphics[width=\columnwidth]{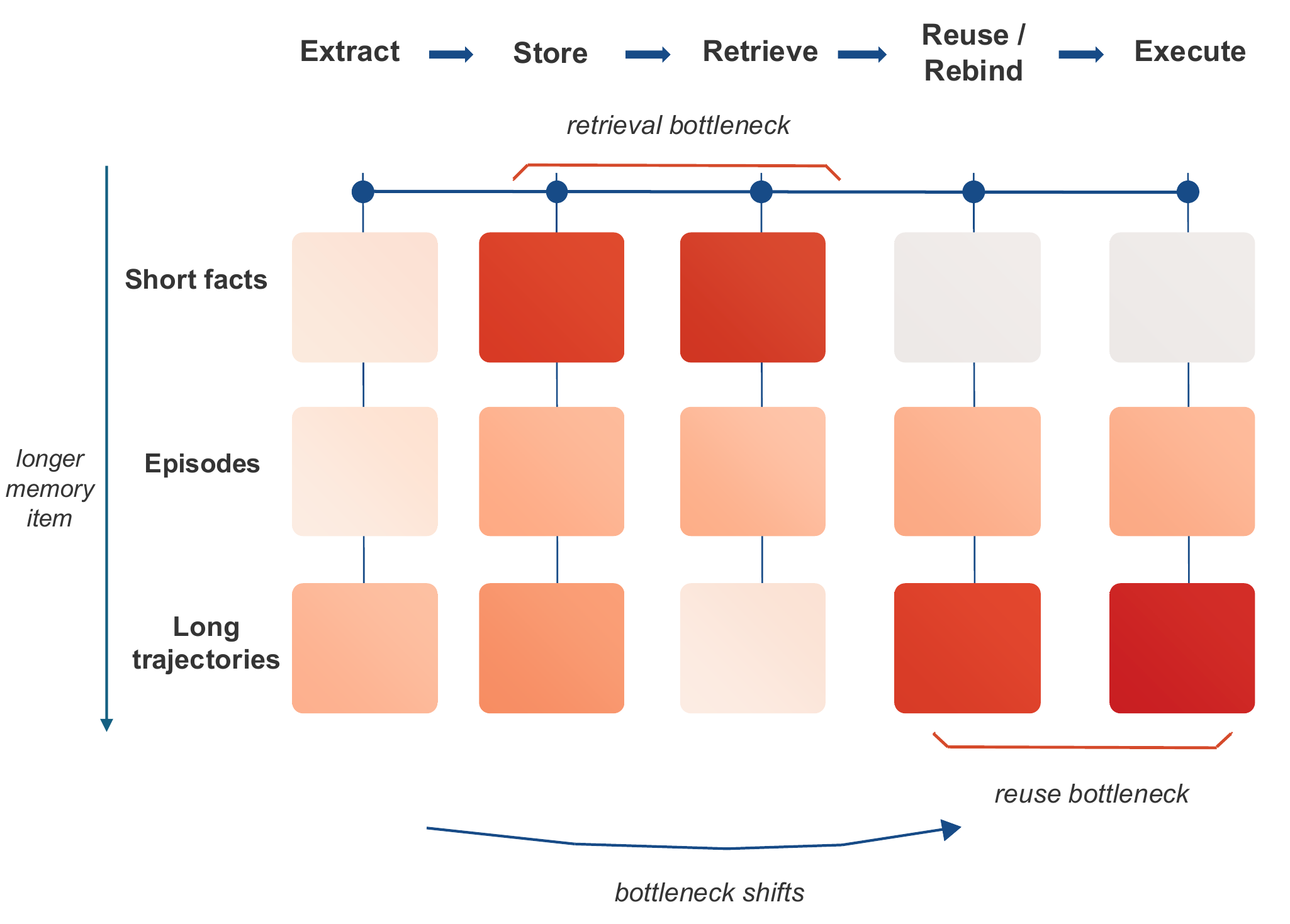}
\caption{\textbf{A design hypothesis: the bottleneck shifts with memory item
length.} For short facts or episodes, retrieving the relevant item often
supplies what the target needs. Long trajectories can save more exploration,
but their target-side value depends on reuse, rebinding, and execution.
\method addresses this post-retrieval step.}
\label{fig:overview}
\end{figure}
Agent memory has progressed from carrying a short interaction history to
maintaining external stores, structured records, retrieval indices, and learned
memory operations. These systems aim to let an agent bring past experience into
a later decision rather than rediscovering the same information or procedure
from scratch. Recent methods make histories easier to retain, organize, and
retrieve
\citep{packer2023memgpt,zhong2024memorybank,gutierrez2024hipporag,
chhikara2025mem0,xu2025amem,hu2026sam,yu2026agenticmemory,li2026ocrmemory},
while long-context benchmarks test whether systems recover evidence over
extended interactions \citep{bai2024longbench,maharana2024locomo,
wu2025longmemeval,tan2025membench,hu2025memoryagentbench}. This progress makes
historical experience accessible. It does not yet establish that the experience
helps an agent solve its current query. This distinction matters because many
memory evaluations naturally end at access: can a system retain an item, rank
it for a query, or answer a question about an earlier interaction? LongBench,
LoCoMo, and LongMemEval make these access questions measurable across long
contexts and conversations \citep{bai2024longbench,maharana2024locomo,
wu2025longmemeval}. They are necessary tests, but they leave open a second
question for an acting agent: once an experience has entered the context, does
it improve the new task that the agent must now complete?

For short, self-contained memory items, retrieval and reuse are often nearly
the same operation. If a query needs a fact, a local instruction, or a compact
episode, returning the right item usually supplies the evidence needed for the
next answer or action. Retrieval quality is therefore a useful proxy for memory
utility in these settings. Long-horizon task experience has a different
structure. A successful trajectory may encode a valuable tool workflow,
decision rule, and verification sequence, so it can save far more exploration
than a short fact. Yet it also carries source-specific users, objects, paths,
dates, observations, and failed branches. Finding such a trajectory does not
tell the agent which part transfers, which binding has expired, or which checks
must be repeated before it acts. This is the regime targeted by interactive
agent environments: browser, API, and database tasks couple many actions to a
changing state and evaluate their final consequences
\citep{zhou2024webarena,drouin2024workarena,trivedi2024appworld,
yao2025taubench}. A literal trace can place the agent in the right subsystem
while still applying an obsolete argument to the wrong current object.

Figure~\ref{fig:overview} expresses the resulting bottleneck shift. Moving
from short facts through episodes to long trajectories, a memory item can carry
more of the work that an agent would otherwise repeat. The main difficulty also
moves rightward along the memory pipeline. Once a relevant long trajectory has
been found, the target agent must extract the procedure that still applies,
recover current bindings, reject source details that no longer hold, and verify
the new final state. Retrieval remains necessary, but it is no longer enough.
The difficulty is also not merely a context-window problem. Giving an actor a
long raw trace can bury a current objective beneath old observations and
incidental branches, a failure consistent with evidence that models do not
always use the relevant portion of a long context reliably
\citep{liu2024lostmiddle}. Experience-memory methods accordingly extract
reflections, skills, workflows, or reusable knowledge from prior runs
\citep{shinn2023reflexion,zhao2024expel,wang2024awm}; what remains unclear is
how to assess the value of that extracted support for a later query.

This observation changes the basic evaluation question. Memory should be judged
by whether past experience improves the solution of the current query: higher
verified completion, less unnecessary exploration, or lower online cost without
discarding necessary checks. A high retrieval score alone cannot answer this
question for long trajectories. Consider an earlier multi-app workflow for
search, verification, and artifact creation. A later request may preserve that
workflow while changing the person, file, date, or environment state. Starting
from scratch wastes the earlier run; replaying it can copy an obsolete
recipient, path, or state assumption. The useful support is a target-bound
account of the procedure, the bindings to recover, and the checks that remain
necessary. We call this operation query-conditioned reuse.

This criterion separates memory utility from both retrieval quality and source
fidelity. A support object may faithfully preserve a source trajectory yet
harm the target by carrying stale bindings forward; conversely, an aggressively
short summary may save tokens while omitting the precondition that prevents an
invalid action. The relevant comparison therefore holds the available history
fixed and asks which use of that history gives the target agent the best
verified outcome for its own environment.

We introduce an end-to-end setting that evaluates this post-retrieval operation
directly. A unified frozen bank contains verified historical trajectories. For
each target, the retriever returns candidate experiences and a shared ranker
selects one record before any memory condition runs. Full Trajectory, Generic
Summary, and \method therefore receive the same selected experience but use it
differently. Target Success, Milestone completion, API calls, and online tokens
then measure whether that experience actually helps the current query. Our
contribution is a problem formulation and evaluation protocol for this boundary,
together with \method, a minimal target-conditioned support transformation.
The analyses test how selected-memory length and source--target binding shift
change the utility of direct reuse.

\section{Related Work}
\subsection{Long-Context and Retrieval-Oriented Memory}
Memory systems study storage, indexing, updating, and context delivery,
including MemGPT, MemoryBank, HippoRAG, Mem0, and A-MEM
\citep{packer2023memgpt,zhong2024memorybank,gutierrez2024hipporag,
chhikara2025mem0,xu2025amem}. LongMemEval, LongBench, LoCoMo, MemBench, and
MemoryAgentBench evaluate retention and access over long or incremental
interactions \citep{wu2025longmemeval,bai2024longbench,maharana2024locomo,
tan2025membench,hu2025memoryagentbench}; RAG supplies the standard
retrieve-then-condition pattern \citep{lewis2020rag}, and surveys organize
memory operations and representations \citep{zhang2024memorysurvey,
du2025rethinkingmemory}. Transformer-XL and Memorizing Transformers make much
longer access possible \citep{dai2019transformerxl,wu2022memorizing}, but more
context does not guarantee that a model uses the relevant portion
\citep{liu2024lostmiddle}. We therefore ask a downstream question: after a
trajectory has been selected, what support lets an agent use it on a new task?

\subsection{Trajectory and Procedural Experience}
Prior experience can appear as reflection, a skill, a script, or a retrieved
trajectory: Reflexion, Generative Agents, Voyager, ReAct, ExpeL, Synapse, and
Agent Workflow Memory instantiate these choices
\citep{shinn2023reflexion,park2023generative,wang2023voyager,yao2023react,
zhao2024expel,zheng2024synapse,wang2024awm}. SAM uses state-adaptive cues,
Agentic Memory learns memory operations, and OCR-Memory trades representation
for faithful long-history access \citep{hu2026sam,yu2026agenticmemory,
li2026ocrmemory}. We instead fix the candidate set and selected trajectory,
then measure whether its representation changes a later target's outcome or
online cost.

\subsection{Long-Horizon Agent Evaluation}
AgentBench, AppWorld, WebArena, WorkArena, and $\tau$-bench provide interactive
and verifiable settings for multi-step agency
\citep{liu2024agentbench,trivedi2024appworld,zhou2024webarena,
drouin2024workarena,yao2025taubench}. Mind2Web, AndroidWorld, WebVoyager,
GAIA, OSWorld, and SWE-bench broaden this coverage across web, mobile,
multimodal, and software tasks \citep{deng2023mind2web,rawles2025androidworld,
he2024webvoyager,mialon2024gaia,xie2024osworld,jimenez2024swebench}. These
benchmarks usually score isolated execution. Our unified bank instead lets a
target search same- and cross-environment history while retaining its native
verifier, so the outcome measures whether prior experience reduces target work
without becoming an identical replay.

\section{Task Setting: Query-Conditioned Trajectory Reuse}
\begin{figure*}[t]
\centering
\includegraphics[width=\textwidth]{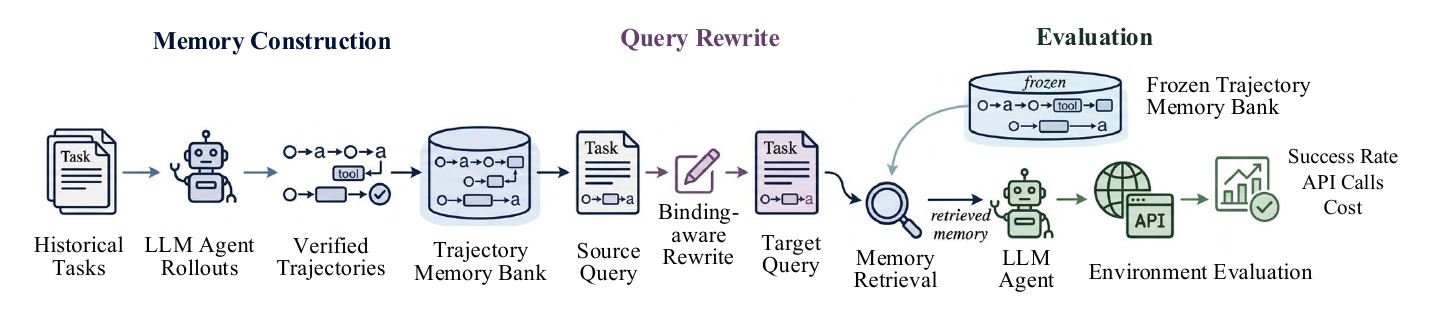}
\caption{\textbf{Evaluation pipeline for query-conditioned trajectory reuse.}
Offline, verified historical rollouts populate one unified memory bank. A
binding-aware rewrite creates a target query with a related workflow
but new target-specific values. Online, a fixed retriever returns histories
from the frozen bank, after which the agent executes only against the target
environment and its own verifier. The comparison measures target Success,
Milestone completion, API calls, and non-overlapping online tokens.}
\label{fig:task-setting}
\end{figure*}

Figure~\ref{fig:task-setting} defines the evaluation unit used in this paper.
The unit is a \emph{source--target pair}: a verified historical trajectory
and a later task that preserves a reusable procedure while changing the values
needed to carry it out. The source is not a demonstration to replay. It is a
record of one successful interaction that may contain a procedure, failed
branches, and state checks. The target is a new task with its own initial
state, tool feedback, and verifier. Memory helps only when the agent extracts
the part of the source that still applies and re-obtains the values that no
longer do.

\subsection{Problem Definition}
For a target task $t$, let $q_t$ denote its natural-language query and let
$o_{t,0}$ be the initial observation. Before the target starts, the agent has
access to a frozen bank $\mathcal{B}$ of verified historical trajectories. A
fixed retriever $R$ returns the same top-$k$ records for every compared method,
\[
Z_t = R(q_t,o_{t,0},\mathcal{B}).
\]
Given $Z_t$, the target query, and the initial observation, a reuse mechanism
$\rho$ writes a support object $r_t=\rho(Z_t,q_t,o_{t,0})$. The acting agent
then produces a new target trajectory $\hat{\tau}_t$ from
$(q_t,o_{t,0},r_t)$ and receives the target's own verifier outcome. This setup
separates two questions that are often conflated: whether the bank retrieves a
potentially relevant history, and whether the agent can turn that history into
an action plan for the current task.

The target evaluation rewards verified completion and penalizes online work.
We therefore report success or verifier score together with API calls and
token cost. A longer or more literal memory representation is not preferred by
definition; it must reduce the work of solving the target. All methods start
from the same cached $Z_t$ and the same ranker-selected record, so a difference
cannot arise from retrieval or source selection. They differ only in the
representation and use of that selected experience.

\subsection{Offline Memory Construction}
The bank stores episodes rather than author-written skills or task summaries.
We construct one unified frozen bank of 623 verified historical trajectories
from successful source-task executions across WebArena, WorkArena, and
AppWorld. An agent solves each source task through its native interface, and a
rollout enters $\mathcal{B}$ only after that environment's checker accepts its
final state. We jointly index all trajectories in one mixed memory pool rather
than partitioning records by task family or environment. The supplementary
material reports the bank's source-benchmark composition and complete manifest.

Each retained record contains the source instruction, the ordered sequence of
observations and actions, tool calls with their arguments and returned
observations, terminal artifacts when present, and the verifier result. We
retain the environment, task identifier, and rollout configuration for audits,
but exclude these provenance fields from retrieval representations and target
prompts. The bank can therefore contain detours and failed attempts that
occurred before a successful final state, as it would in a deployed agent
system.

We construct the bank from task families of intermediate difficulty. A
no-memory baseline is run repeatedly on candidate tasks, and we retain families
that yield both verified successes and verified failures. Trivial tasks leave
little room for prior experience to matter, whereas tasks with no successful
rollouts provide no verified trajectory to store. Successful source runs from
the retained families form the frozen bank used at evaluation time.

\subsection{Binding-Aware Target Construction}
Starting from each verified source trajectory $\tau_s$, we create up to four
target variants by retaining the source's intended workflow while replacing
one or more target-specific bindings at different divergence levels. A binding
is a value that an agent must ground in the current task or environment, such
as an entity, a user, a record identifier, a file or location, a date, a
parameter, or the relevant current state. The rewrite may preserve a procedure
such as ``inspect, validate, modify, and verify,'' but it never licenses the
agent to copy the source values into the target. Some trajectories cannot
support every rewrite pattern because of task-specific constraints. The final
benchmark therefore contains 2,391 valid target task instances, or 3.84 target
variants per historical trajectory on average. The supplementary material
reports target counts by benchmark and selected-memory-length group.

The target starts from its own state and is checked by its own executable
verifier or pre-specified rubric. Literal replay of $\tau_s$ can therefore
fail even when its workflow remains useful; the agent must discover the
current bindings through target-side observations and tool calls. We retain the
source--target relation only as sealed audit metadata for relevance annotation
and error analysis. It does not appear in bank records, retrieval indices,
prompts, or target-agent inputs.

\subsection{Frozen Retrieval and Evaluation Boundary}
After bank construction, all verified source episodes are pooled into one
snapshot and frozen. The retriever indexes only visible source instructions
and trajectory descriptors, not environment labels, task identities, sealed
source--target relations, verifier diagnostics, or target-conditioned
summaries. Given $q_t$ and $o_{t,0}$, a fixed embedding retriever produces a
top-$5$ candidate set $Z_t$. A lightweight ranking stage then selects one
trajectory from $Z_t$; we cache both the candidate set and the selected record
before target execution and reuse them for Full Trajectory, Generic Summary,
and \method\ conditions. The retriever may return a transferable workflow, a
superficial match, or nothing useful. No condition receives an oracle source
trajectory.

All compared conditions share the target query and state, frozen bank, cached
$Z_t$, ranker-selected trajectory, acting model, decoding configuration, and
tool budget. \method{} may read only the selected trajectory, $q_t$, and
$o_{t,0}$ while writing $r_t$; it cannot call the environment privately or
replace the cached selection. In partially observable settings, the acting
agent performs any required target-side discovery and pays the resulting cost.

For each target run, let $I_{\mathrm{base}}$ be the non-memory acting prompt,
$I_{\mathrm{mem}}$ the retrieved context shown to the acting agent,
$I_{\mathrm{syn}}$ and $O_{\mathrm{syn}}$ the support-synthesis tokens, and
$O_{\mathrm{act}}$ the acting-agent output. We report
\[
C_{\mathrm{online}} =
I_{\mathrm{base}}+I_{\mathrm{mem}}+I_{\mathrm{syn}}+
O_{\mathrm{syn}}+O_{\mathrm{act}},
\]
alongside API calls. The API-reported acting input is
$I_{\mathrm{base}}+I_{\mathrm{mem}}$, so we report it as a breakdown rather
than add it twice. Source-rollout cost belongs to offline bank construction and
is logged separately.

Because a target can have more than one useful predecessor, we audit retrieval
without assuming a single oracle source. Blind annotators label each returned
record as irrelevant, surface-related but procedurally unusable,
workflow-relevant, or highly actionable. We report
\emph{usable-memory@1}, \emph{usable-memory@k}, and the rank of the first
workflow-relevant record. These diagnostics tell us whether retrieval exposes
history that could help; the end-to-end metrics determine whether the agent
actually converts that opportunity into a successful, efficient target run.

\section{A Minimal Query-Conditioned Reuse Framework}
\subsection{Design Principle}
The framework is intentionally simple. It does not replace the memory store,
retriever, or acting agent. Instead, it inserts one operation after retrieval:
produce compact reuse support explicitly conditioned on the target query and
current state. This makes the framework a diagnostic intervention. If it helps
after the same retrieved records have been fixed, the improvement supports the
claim that the missing operation is reuse rather than storage alone.

\subsection{Reusable Support}
Given $Z_t$, $q_t$, and $o_{t,0}$, \method{} produces a short support object
with four fields:
(i) a workflow invariant, (ii) bindings that must be re-obtained in the target,
(iii) applicability conditions (including when to decline reuse), and
(iv) a verification guardrail. These fields are a minimal implementation
choice, not a claim that one universal memory schema is optimal. The support
must be substantially shorter than the retrieved set and must not
reveal target answers or hidden evaluator information.

The workflow field keeps only the action pattern that the target still needs:
for example, inspect the current object, verify the relevant condition, carry
out a modification, and validate the result. The re-obtain field blocks a
common misuse of trajectory memory. A source trace can mention a user name,
file path, account, object identifier, or prior artifact that helped solve the
source task but says nothing about the target value. The reuse note names that
dependency without supplying the old value as an answer. Applicability and
verification fields retain the reason an earlier agent paused, changed branch,
or validated. They make non-reuse a valid outcome when target state violates a
source precondition, rather than encouraging an agent to replay history.

This representation deliberately stays small. A learned graph, a hierarchy of
summaries, or a new persistent store may outperform it later. They would not,
by themselves, establish whether the useful operation lies before or after
retrieval. The minimal design keeps the intervention legible: given the same
retrieved records, the target agent receives either the raw history or a
target-conditioned account of how to use it.

\subsection{Target Execution}
The acting agent receives either no memory, the selected raw trajectory, its
generic source-only summary, or the \method support object. A lightweight
ranker uses compact descriptors of the cached top-$5$ records and the target
query to select one record before any target condition runs. Generic summaries
are produced offline from each historical record alone, before the target query
arrives; their length budget matches the \method support budget. Thus the
memory conditions share the same $Z_t$, selected trajectory, target query,
initial target observation, tool access, and target-side generation budget.
They differ only in how the same selected experience is represented and used.
In the evaluation, the acting model, decoding settings, tool
budget, and target rollouts are held fixed within each target across
conditions. We report the cost of producing \method support separately and in
the online reuse total. This avoids treating a reduction in target-agent output
alone as a free gain.

We use DeepSeek-V4-Pro \citep{deepseek2026v4pro} for both historical and target
runs. Prompts state that historical information is advisory rather than an
instruction to replay actions literally. A raw-trajectory baseline can inspect
and exploit its history, whereas \method receives no extra tool privileges,
target state, or verifier hints. The comparison changes only the use of the
same selected historical experience: raw delivery, source-only compression, or
target-bound support construction.

\section{Experiments}
\input{experiment-results}

\section{Discussion and Limitations}
The experiments isolate the value of a verified prior trajectory after it has
entered the memory pipeline. Candidate selection is already strong: reranking
selects reusable history for 94.8\% of targets and trails oracle reusable
selection by 1.8 success points. The results are consistent with a remaining
post-selection cost when long histories and changed bindings expose the actor
to source details that no longer apply.
This distinction matters for memory-system design. A store may preserve a
complete record for evidence and provenance, while the acting prompt should
contain a compact, target-bound account of the reusable procedure.

The study has a narrow boundary. It evaluates successful source trajectories,
a single selected memory, and controlled source--target binding shifts; it does
not measure naturally recurring task histories, partial failures, multi-memory
composition, or open-ended memory acquisition. Those settings may change both
the available procedures and the state that an agent must recover. The reported
token savings also do not mean that every task should use fewer tokens:
safety-sensitive targets can require additional checks. We measure verified
completion, but not irreversible side effects or policy violations caused by a
reused trajectory. We therefore treat cost as one outcome beside verified
completion, rather than as a goal on its own.

The comparison holds the retrieved candidate set, acting model, decoding,
tool budget, and target-state access fixed. It attributes differences to the
representation and use of the same selected trajectory; it does not by itself
isolate every field of the support schema. Future work can replace the
embedding retriever or learn a reuse policy, but it should retain this
accounting boundary and test whether the resulting help reduces target work
without importing stale source bindings.

\section{Conclusion}
We study agent memory by asking how past experience helps a current query.
Long completed trajectories help, but raw delivery and a generic source-only
summary leave useful target-specific work unresolved. Given the same selected
historical trajectory, \method raises success to 62.3\% and reduces online
tokens by 48.9\% relative to Full Trajectory. The selection analysis shows that
a reusable memory is available for 94.8\% of targets after ranking. As
histories become longer or target bindings move farther from the source,
raw-trajectory utility falls, while target-bound support retains a larger
measured gain.

Memory systems should therefore preserve rich records in storage while giving
the actor a compact, target-bound account of the procedure, the bindings it
must recover, and the checks that still apply. The setting leaves room for
better retrievers and learned reuse policies, but it makes their test clear:
they must improve target success without hiding the cost of the help.

This framing also changes how memory baselines should be interpreted. A raw
trajectory is not simply a stronger version of a short summary because it
contains more tokens; it is an intervention that exposes an actor to both
useful procedure and obsolete state. Conversely, a short description is not
automatically useful merely because it is cheap. The relevant question is
whether the information sent after retrieval lets the target agent take fewer
unnecessary actions while still checking the values that changed. By fixing
candidate retrieval, target state, model, decoding, and tool budget, the
present comparison evaluates that question after candidate retrieval has
ended. Future systems can use different stores, retrievers, or learned support
writers, but should report the same distinction between what was retrieved,
what was selected, what was delivered to the actor, and what the actor verified
in the target environment.

\bibliography{references}

\input{appendix}
\end{document}

%% file: experiment-results.tex
\subsection{Protocol and Metrics}
We compare four conditions under the frozen-retrieval protocol in
Section~3: \textsc{No Memory}, \textsc{Generic Summary},
\textsc{Full Trajectory}, and \method. For every target, an embedding retriever
returns the same top-$5$ historical trajectories for the three memory
conditions. A lightweight ranker selects one trajectory from that shared set
before any condition runs. Full Trajectory supplies the selected record
directly, Generic Summary supplies its source-only summary, and \method writes
its query-conditioned support object from the same selected record. Thus, the
acting agent never receives five long trajectories at once, and
Table~\ref{tab:overall-results} isolates how the selected experience is
represented and used. The selection diagnostic below evaluates the ranker
separately.

We report verified success, milestone completion, API calls, and non-overlapping
online tokens. Online tokens include the acting prompt, selected memory,
support-synthesis input and output, and acting output. To measure reuse rather
than task difficulty alone, the stratified analyses report
\[
U = S_{\mathrm{memory}} - S_{\mathrm{no\ memory}}.
\]
Candidate relevance and final-memory relevance were judged against the sealed
source--target relation and a reusable-workflow annotation. A paired trajectory
is the historical trajectory used to construct the target; a reusable trajectory
may be another history if it supplies a valid procedure for the target.

\subsection{Benchmarks, Model, and Controls}
We evaluate end-to-end target execution in WebArena
\citep{zhou2024webarena}, WorkArena \citep{drouin2024workarena}, and AppWorld
\citep{trivedi2024appworld}. Each environment supplies its own initial state,
tool interface, and success or milestone checker, while retrieval searches the
same unified mixed memory bank across environment boundaries. We use DeepSeek-V4-Pro
\citep{deepseek2026v4pro}
for source rollouts, summary ranking, support synthesis, and target execution.
Within a target, every condition shares the target suite, initial state,
cached retrieval result, ranker-selected trajectory, model, decoding
configuration, tool budget, and verifier. The comparison therefore changes
only the representation and use of the selected experience after retrieval.

\subsection{Memory Representations and Run Records}
\textsc{No Memory} receives no historical record. \textsc{Generic Summary}
receives a source-only summary prepared before the target query arrives, so it
cannot encode target answers or current environment observations.
\textsc{Full Trajectory} receives the ranker-selected historical record with
its ordered observations, actions, tool arguments, and returned outputs. A
common ranker sees compact descriptors of the top-$5$ candidates and selects
one source trajectory for every memory condition. \method then writes four
short fields from that selected trajectory: the workflow invariant, bindings to
re-obtain, applicability conditions, and a verification guardrail. The acting
agent receives that note rather than the raw trajectory or an oracle
source--target mapping.

For each run, we log the cached candidate identifiers, selected memory,
verifier outcome, milestone score, API calls, and non-overlapping token
components. Online-token accounting includes the base acting prompt, delivered
memory, support-synthesis input and output, and acting output; it excludes the
offline cost of collecting verifier-approved source trajectories. These records
make the reported efficiency comparison traceable to the same target run
rather than to different retrieval outcomes.

The supplement supplies the prompt template, configuration tables, annotation
definitions, and audit-ledger schema needed to interpret the reported tables
and diagnostics.

\subsection{End-to-End Performance}
\begin{table*}[!t]
\centering
{\small
\setlength{\tabcolsep}{3.5pt}
\begin{tabular}{lcc|cc|cc|cc}
\hline
& \multicolumn{2}{c|}{WebArena} & \multicolumn{2}{c|}{WorkArena} &
\multicolumn{2}{c|}{AppWorld} &
\multicolumn{2}{c}{Efficiency} \\
\cline{2-9}
Method & Success $\uparrow$ & Milestone $\uparrow$ & Success $\uparrow$ & Milestone $\uparrow$ &
Success $\uparrow$ & Milestone $\uparrow$ &
API Calls $\downarrow$ & Online Tokens $\downarrow$ \\
\hline
No Memory & 31.5 & 47.3 & 36.6 & 52.8 & 47.1 & 61.0 & 24.6 & 15.2k \\
Generic Summary & 40.2 & 55.8 & 45.9 & 61.7 & 57.6 & 70.0 & 20.8 & \textbf{8.1k} \\
Full Trajectory & 43.8 & 61.2 & 49.6 & 66.5 & 61.4 & 72.7 & 21.9 & 18.4k \\
\method & \textbf{54.7} & \textbf{70.6} & \textbf{60.4} & \textbf{74.8} &
\textbf{71.8} & \textbf{82.9} & \textbf{16.7} & 9.4k \\
\hline
\end{tabular}
}
\caption{\textbf{End-to-end performance across 2,391 target instances.} Success and
Milestone are percentages; API Calls and Online Tokens are means.}
\label{tab:overall-results}
\end{table*}

Table~\ref{tab:overall-results} shows that historical experience helps, but
the procedure used after candidate retrieval matters. Generic Summary gains
9.5 success points over No Memory, while Full Trajectory gains a further 3.7
points. \method reaches 62.3\% success, 10.7 points above Full Trajectory,
with the fewest API calls among the memory conditions. Its 9.4k online tokens
are about half of the 18.4k required by direct trajectory injection. The
result therefore does not come from sending the actor more historical context.
The same ranking holds in WebArena, WorkArena, and AppWorld: \method is best
on both Success and Milestone in all six environment-specific comparisons.
Relative to Full Trajectory, its Success margin is 10.9 points in WebArena,
10.8 in WorkArena, and 10.4 in AppWorld. The corresponding gains over No
Memory are 23.2, 23.8, and 24.7 points. This consistency matters because the
three environments differ in interaction modality and state observability; the
effect is not carried by one easier benchmark.

\input{selection-results}

\begin{figure*}[!t]
\centering
\includegraphics[width=0.96\textwidth]{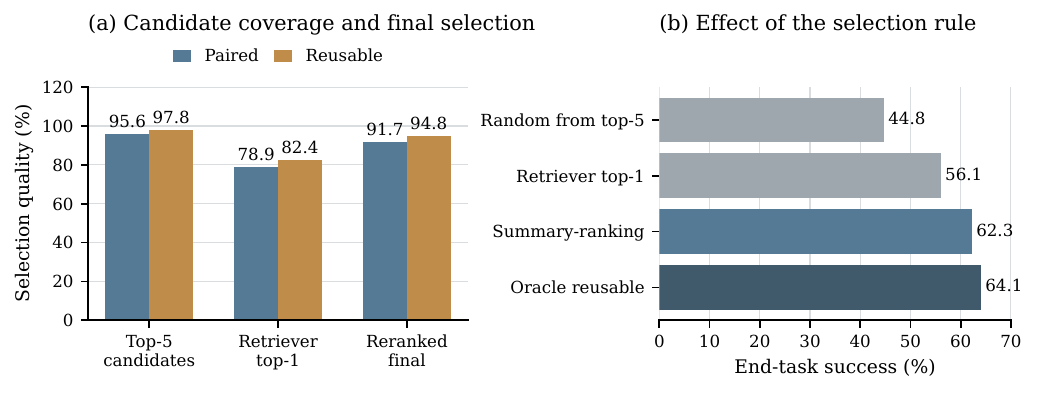}
\caption{Why summary reranking matters. Left: the top-$5$ candidate
set has high paired and reusable coverage, while direct top-$1$ retrieval is
less reliable. Reranking compact candidate summaries restores final-memory
quality without presenting five full trajectories to the acting agent. Right:
the resulting end-task success nearly matches oracle reusable selection.}
\label{fig:selection-quality}
\end{figure*}

\subsection{Sensitivity to Selected-Memory Length}
We partition target instances by the effective-action length of the
ranker-selected memory trajectory: Short has 5--10 actions, Medium 11--20,
Long 21--35, and Very Long more than 35. The same selected memory defines a
length group for every compared condition, including No Memory. No-Memory
Success falls from 55.2\% in the Short group to 18.9\% in the Very Long group.
The groups therefore differ substantially in task difficulty, so
Table~\ref{tab:length-utility} reports within-group utility rather than raw
success.

Direct injection degrades steeply: Full Trajectory falls from +18.4 points for
short histories to +2.9 for very long ones. Generic Summary retains a little
more of its initially smaller gain, but its utility never reaches that of
\method. Query-conditioned support also becomes less useful as histories
lengthen, yet it retains +13.2 points for very long trajectories and 60.3\% of
its short-trajectory utility. Full Trajectory retains only 15.8\% of its
short-trajectory utility; Generic Summary retains 32.4\%. Because the length
groups also differ in no-memory difficulty, this is an association under the
registered construction rather than a causal estimate of length alone.

\begin{table}[!t]
\centering
{\small
\setlength{\tabcolsep}{1.5pt}
\begin{tabular*}{\columnwidth}{@{\extracolsep{\fill}}lccc}
\hline
Length group & Full Trajectory & Generic Summary & \method \\
\hline
Short & +18.4 & +14.2 & +21.9 \\
Medium & +14.1 & +11.3 & +20.4 \\
Long & +8.5 & +7.1 & +17.6 \\
Very Long & +2.9 & +4.6 & +13.2 \\
\hline
\end{tabular*}
}
\caption{\textbf{Memory utility by selected-memory trajectory length.} Entries are
percentage-point gains over No Memory within each length group.}
\label{tab:length-utility}
\end{table}

\subsection{Binding Shift Is Associated with the Reuse Gap}
We next vary the number and type of target-specific bindings rewritten from
the source task. A small rewrite changes one local binding; a medium rewrite
changes two or three bindings or one central constraint; a large rewrite
changes at least four bindings, or both the target entity and initial
environment state. The no-rewrite condition preserves the original intent and
tests same-intent recovery.

Table~\ref{tab:rewrite-utility} identifies the failure mode suggested by the
main result. When no binding changes, Full Trajectory has high utility
(+26.9). Under a large rewrite, its utility shrinks to +2.2, and the generic
summary reaches only +5.3. \method declines as the target moves further from
the source, but preserves +20.1 points under the largest shift. The gain
comes with fewer stale-binding errors: at large shift, direct trajectories
produce stale bindings on 46.9\% of targets, compared with 10.9\% for
\method, while correct rebinding rises from 31.7\% to 77.8\%. We count a
stale-binding error when an action, output, or tool argument repeats a
source-specific value that conflicts with the target query or target-side
observation.
Under large shift, Full Trajectory retains 8.2\% of its no-shift utility
(2.2/26.9), whereas \method retains 67.9\% (20.1/29.6). The method does not
make binding shift disappear; it reduces the rate at which stale source values
displace current-task evidence.

\begin{table}[!t]
\centering
{\small
\setlength{\tabcolsep}{1.5pt}
\begin{tabular*}{\columnwidth}{@{\extracolsep{\fill}}lccc}
\hline
Binding shift & Full Trajectory & Generic Summary & \method \\
\hline
None & +26.9 & +18.2 & +29.6 \\
Small & +19.3 & +15.6 & +28.6 \\
Medium & +9.7 & +10.2 & +24.5 \\
Large & +2.2 & +5.3 & +20.1 \\
\hline
\end{tabular*}
}
\caption{\textbf{Memory utility under binding shift.} Entries are percentage-point
gains over No Memory within each rewrite level.}
\label{tab:rewrite-utility}
\end{table}

\subsection{Interpretation}
The four results form a consistent account. Historical trajectories offer
useful procedural information, since both memory baselines beat No Memory.
Candidate retrieval and single-memory selection are accurate enough that a
substantial part of the remaining loss occurs after a relevant trajectory
reaches the actor. The length and rewrite analyses associate weaker direct
reuse with long source traces and larger target differences. \method keeps the
workflow while requiring the actor to recover current bindings, a mechanism
consistent with its higher success at lower online cost.

Selection and reuse are separate stages: a ranker decides whether usable
history reaches the actor, while the delivered representation determines
whether it can be applied without copying source-side values.

%% file: selection-results.tex
\subsection{Retrieval and Single-Memory Selection}
Figure~\ref{fig:selection-quality} separates candidate coverage from the
decision about what to inject. The embedding retriever places the paired
trajectory in the top five for 95.6\% of targets and at least one reusable
trajectory for 97.8\%. Its top-one paired accuracy, however, is only 78.9\%.
Ranking candidate summaries against the target raises final paired-memory
accuracy to 91.7\% and final reusable-memory accuracy to 94.8\%; only 5.2\%
of selected memories are irrelevant. The mean reciprocal rank of the paired
trajectory is 0.87. Relative to direct top-$1$ retrieval, reranking gains
12.8 points in paired accuracy and 12.4 points in reusable-memory accuracy.

The selection ablation gives the same picture at the task level. Directly
using the retriever's first item lowers success to 56.1\%, and selecting a
random top-five item lowers it to 44.8\%. The ranking prompt reaches 62.3\%,
only 1.8 points below an oracle that selects a reusable candidate. Candidate
selection still leaves headroom, but it is not the main source of end-task
failure in this setting. Figure~\ref{fig:selection-quality} visualizes both
parts of this result: broad top-$5$ coverage enables reranking, and the
reranked choice closes most of the gap to oracle end-task success. The
6.2-point improvement over retriever top-$1$ shows that choosing the memory,
rather than increasing the number of injected trajectories, accounts for the
gain.

%% file: appendix.tex

\appendix
\setcounter{table}{0}
\renewcommand{\thetable}{A\arabic{table}}

\section{Task Construction and Retrieval Protocol}
\label{app:protocol}

The study uses a unified bank of verified historical trajectories. For each
target, retrieval returns a shared top-$5$ candidate set, and the ranker
selects one trajectory before any memory condition runs. The target state,
model, decoding settings, tool budget, random seed, and verifier remain fixed
across conditions. Candidates may come from any of the three environments,
but environment labels are excluded from the retrieval representation.

\begin{center}
\centering
\small
\resizebox{\columnwidth}{!}{%
\begin{tabular}{ll}
\hline
Statistic & Value \\
\hline
Average target variants per source & 3.84 \\
Memory bank & Unified across all environments \\
Retrieval scope & One pooled bank; no environment labels in retrieval input \\
Top-$k$ retrieval & 5 \\
Selected memories per target & 1 \\
Random seeds & 3 seed-matched runs per target and condition \\
\hline
\end{tabular}}
\captionof{table}{Memory-bank and evaluation settings.}
\label{tab:app-memory-settings}
\end{center}

\begin{center}
\centering
\small
\resizebox{\columnwidth}{!}{%
\begin{tabular}{lrrrr}
\hline
Environment & Sources & Targets & Paired targets & Seeds \\
\hline
WebArena & 228 & 874 & 874 & 3 \\
WorkArena & 201 & 772 & 772 & 3 \\
AppWorld & 194 & 745 & 745 & 3 \\
\hline
Total & 623 & 2,391 & 2,391 & 3 \\
\hline
\end{tabular}}
\captionof{table}{Source-trajectory and target-task inventory after exclusions.
Target counts enumerate unique target instances, not seed-expanded runs. Each
reported result averages the three seed-matched runs. ``Paired target'' denotes
the target constructed from the listed source trajectory; it does not identify
the record ultimately selected by the ranker.}
\label{tab:app-task-inventory}
\end{center}

\subsection{Binding-Shift Construction}

The no-rewrite condition preserves the source intent. Small, medium, and large
rewrites change one local binding, two or three bindings or one central
constraint, and at least four bindings or both the target entity and initial
environment state, respectively. The rewrite procedure preserves the task
family while requiring the acting agent to ground values in the target.

\begin{center}
\centering
\small
\begin{tabular}{lr}
\hline
Rewrite level & Number of targets \\
\hline
None & 623 \\
Small & 602 \\
Medium & 588 \\
Large & 578 \\
\hline
Total & 2,391 \\
\hline
\end{tabular}
\captionof{table}{Distribution of source--target binding rewrites.}
\label{tab:app-rewrite-distribution}
\end{center}

\begin{center}
\centering
\small
\resizebox{\columnwidth}{!}{%
\begin{tabular}{ll}
\hline
Item & Setting \\
\hline
Embedding retriever & BGE-M3 \\
Retrieval unit & One complete trajectory \\
Retrieval scope & Unified cross-environment bank \\
Candidate size & Top-5 \\
Ranker & DeepSeek-V4-Pro \\
Duplicate handling & Remove near-duplicate trajectories before indexing \\
Tie-breaking & Highest reranking score \\
\hline
\end{tabular}}
\captionof{table}{Retrieval and reranking configuration. The candidate bank is
pooled across environments, while environment labels are excluded from the
retrieval input.}
\label{tab:app-retrieval-config}
\end{center}

\section{Model Configuration}

All conditions use DeepSeek-V4-Pro for generation and ranking. The source
rollout and target actor sample at temperature $0.2$; all support-writing and
selection steps use deterministic decoding.

\begin{center}
\centering
\scriptsize
\resizebox{\columnwidth}{!}{%
\begin{tabular}{l l c c r l}
\hline
Component & Model & Temperature & Top-$p$ & Max tokens & Decoding \\
\hline
Source rollout & DeepSeek-V4-Pro & 0.2 & 0.95 & 4,096 & Stochastic \\
Summary writer & DeepSeek-V4-Pro & 0 & 1.0 & 1,024 & Deterministic \\
Ranker & DeepSeek-V4-Pro & 0 & 1.0 & 512 & Deterministic \\
QCR writer & DeepSeek-V4-Pro & 0 & 1.0 & 1,024 & Deterministic \\
Actor & DeepSeek-V4-Pro & 0.2 & 0.95 & 4,096 & Stochastic \\
\hline
\end{tabular}
}
\captionof{table}{Model and decoding settings.}
\label{tab:app-model-config}
\end{center}

\section{Annotation and Rebinding Analysis}

A stale-binding error occurs when an action, output, or tool argument repeats
a source-specific value that conflicts with the target query or a target-side
observation. Correct rebinding requires the actor to recover the target value
before using it. Annotators inspect the source record, target task,
target-side observations, and executed action trace rather than inferring a
label from final success alone.

\begin{center}
\centering
\small
\begin{tabular}{lcc}
\hline
Metric, large rewrite & Full Trajectory & \method \\
\hline
Stale-binding error & 46.9\% & 10.9\% \\
Correct rebinding & 31.7\% & 77.8\% \\
\hline
\end{tabular}
\captionof{table}{Rebinding analysis for large binding rewrites.}
\label{tab:app-rebinding}
\end{center}

\begin{center}
\centering
\small
\begin{tabular}{p{0.44\columnwidth}p{0.41\columnwidth}}
\hline
Item & Value \\
\hline
Annotators & 2 \\
Adjudicator & 1 \\
Agreement (Cohen's $\kappa$) & 0.87 \\
Audit unit & One constructed source--target instance \\
Source--target construction and rebinding audits & 2,391 \\
\hline
\end{tabular}
\captionof{table}{Annotation statistics for source--target construction and
rebinding audits. Candidate relevance is labelled at the returned-record level
using the rubric in Table~\ref{tab:app-relevance-labels}; its count is not
included in the 2,391 source--target audit units.}
\label{tab:app-annotation}
\end{center}

\section{QCR Support Construction}

QCR writes a compact target-bound note from the selected trajectory, target
query, and initial target observation. It does not access the environment or
the verifier while preparing the note. The actor must ground every required
target value through the query, current observation, or a later tool call.

\begin{center}
\scriptsize
\begin{tabular}{p{0.22\columnwidth}p{0.27\columnwidth}p{0.28\columnwidth}}
\hline
Field & Source information retained & Target-side requirement \\
\hline
Workflow invariant & Tool order, decision rule, and prior validation step & Keep only the action pattern that applies to the target. \\
Bindings to re-obtain & Entities, paths, dates, users, record identifiers, and parameters & Recover values from current evidence; copy no source binding. \\
Applicability conditions & Source preconditions, constraints, and branch conditions & Decline reuse when a required precondition does not hold. \\
Verification guardrail & The source check that established completion & Verify the target-side result before completion. \\
\hline
\end{tabular}
\captionof{table}{QCR support schema.}
\label{tab:app-qcr-schema}
\end{center}

\subsection{Support-Writer Prompt Template}

The following prompt template specifies the information supplied to the QCR
writer in this study.

\begin{quote}
\footnotesize
You receive one historical trajectory, the current target query, and the
initial target observation. Write a short support note with four labeled
fields: (1) workflow invariant, (2) bindings to re-obtain, (3) applicability
conditions, and (4) verification guardrail. Treat historical identifiers,
paths, users, dates, tool outputs, and environment state as source-side
evidence, not target answers. Do not infer hidden target state, call tools, or
copy a historical binding into the target. If the source procedure does not
apply, state the condition that blocks reuse.
\end{quote}

The actor receives the note as advisory information. It may follow the
workflow only after checking the target-side preconditions, and it must recover
each listed binding before submitting an action that depends on it.

\section{Retrieval Labels and Selection Details}

Blind relevance annotation distinguishes a lexical match from a trajectory
that supplies an executable procedure. The source--target pairing remains
sealed metadata and does not appear in the retrieved record, ranker input, or
acting prompt.

\begin{center}
\small
\begin{tabular}{p{0.28\columnwidth}p{0.61\columnwidth}}
\hline
Label & Decision rule \\
\hline
Irrelevant & The record does not provide a useful procedure for the target. \\
Surface-related but unusable & The record shares words, tools, or a broad task topic, but its procedure does not transfer. \\
Workflow-relevant & The record supplies a procedure that can transfer after target-specific rebinding. \\
Highly actionable & The record supplies a procedure and target-recoverable conditions that can guide immediate execution. \\
\hline
\end{tabular}
\captionof{table}{Relevance labels used for retrieval and selection audits.}
\label{tab:app-relevance-labels}
\end{center}

\begin{center}
\small
\begin{tabular}{p{0.66\columnwidth}r}
\hline
Metric & Value (\%) \\
\hline
Top-5 paired coverage & 95.6 \\
Top-5 reusable coverage & 97.8 \\
Retriever top-1 paired accuracy & 78.9 \\
Final paired-memory accuracy & 91.7 \\
Final reusable-memory accuracy & 94.8 \\
Direct retriever top-1 end-task Success & 56.1 \\
Random top-5 selection end-task Success & 44.8 \\
Ranker-selected memory end-task Success & 62.3 \\
Oracle reusable selection end-task Success & 64.1 \\
\hline
\end{tabular}
\captionof{table}{Retrieval and single-memory selection diagnostics. Coverage
and accuracy entries are proportions of unique target instances. End-task
Success entries average within environment and seed, then take the unweighted
mean across WebArena, WorkArena, and AppWorld.}
\label{tab:app-selection-details}
\end{center}

\section{Stratified Utility Results}

The following tables give the values underlying the length and binding-shift
analyses in the main paper. Each entry reports the percentage-point change in
Success relative to No Memory within the same stratum.

\begin{center}
\small
\resizebox{\columnwidth}{!}{%
\begin{tabular}{lcccc}
\hline
Source length & Action range & Full Trajectory & Generic Summary & \method \\
\hline
Short & 5--10 & +18.4 & +14.2 & +21.9 \\
Medium & 11--20 & +14.1 & +11.3 & +20.4 \\
Long & 21--35 & +8.5 & +7.1 & +17.6 \\
Very Long & $>35$ & +2.9 & +4.6 & +13.2 \\
\hline
\end{tabular}
}
\captionof{table}{Memory utility by selected-memory trajectory length.}
\label{tab:app-length-utility}
\end{center}

\begin{center}
\small
\resizebox{\columnwidth}{!}{%
\begin{tabular}{lccc}
\hline
Binding shift & Full Trajectory & Generic Summary & \method \\
\hline
None & +26.9 & +18.2 & +29.6 \\
Small & +19.3 & +15.6 & +28.6 \\
Medium & +9.7 & +10.2 & +24.5 \\
Large & +2.2 & +5.3 & +20.1 \\
\hline
\end{tabular}
}
\captionof{table}{Memory utility under binding shift.}
\label{tab:app-rewrite-utility}
\end{center}

\section{Per-Run Audit-Ledger Schema}

The evaluation ledger schema specifies one row for each target-condition
execution. It keeps the retrieval boundary auditable: the same candidate set
and selected record must appear across the compared memory conditions for a
given target. The schema also separates a failed execution from an invalid
reuse decision.

\begin{center}
\footnotesize
\begin{tabular}{p{0.29\columnwidth}p{0.55\columnwidth}}
\hline
Field & Purpose \\
\hline
Environment and target identifier & Locate the target's native task suite, verifier, and initial state. \\
Source candidate identifiers & Reconstruct the cached top-$5$ candidate set. \\
Selected trajectory identifier & Check that every memory condition uses the same ranked source record. \\
Memory condition and random seed & Reproduce the target execution setting. \\
Verifier outcome and milestone score & Separate final completion from partial task progress. \\
API calls and token components & Reconstruct online work without counting an API input twice. \\
Execution error and failed action & Identify tool, state, or verifier failures. \\
Stale-binding and correct-rebinding labels & Audit whether source values displaced target-side evidence. \\
\hline
\end{tabular}
\captionof{table}{Fields retained for every target-condition run.}
\label{tab:app-run-ledger}
\end{center}

Source-rollout cost remains outside the online total because it belongs to
offline bank construction. Online accounting includes the non-memory acting
input, delivered memory, QCR-writer input and output, and actor output. The
acting API input contains the first two terms only, so it is recorded as a
breakdown rather than counted again.

\subsection{Information Boundary}

The retriever indexes visible source instructions and trajectory descriptors.
It does not index environment labels, task identities, sealed source--target
relations, verifier diagnostics, or target-conditioned summaries. QCR may read
only the selected source record, target query, and initial target observation;
it cannot query the environment privately, replace the selected record, or see
hidden evaluator information. The acting agent pays for any target-side
discovery through its own tool calls.

\paragraph{Case-study reporting.}
Qualitative examples should show the source procedure and the changed target
binding side by side. A useful pair contains one Full Trajectory failure that
copies a stale value and one QCR success that recovers the value from current
evidence. Private identifiers, credentials, and hidden evaluator material must
be redacted before release.

\begin{center}
\footnotesize
\begin{minipage}{\columnwidth}
\centering
\begin{tabular}{p{0.30\columnwidth}p{0.54\columnwidth}}
\hline
Case element & Material shown to the reader \\
\hline
Source task and selected record & Short task intent and the transferable action sequence. \\
Target task and changed binding & Target intent, with the entity, path, date, or record value that changed. \\
Target-side evidence & The query text, observation, or tool result used to recover the new value. \\
Full Trajectory decision & The copied source value and its verifier or execution consequence. \\
QCR note and final action & The workflow, binding warning, verification guardrail, and target-grounded action. \\
Outcome & Verifier result, milestone result when available, and sanitized failure category. \\
\hline
\end{tabular}
\captionof{table}{Required layout for each released source--target case study.}
\label{tab:app-case-layout}
\end{minipage}
\end{center}